\documentclass{article}
\usepackage[margin=1.5in]{geometry}

\usepackage[utf8]{inputenc} 
\usepackage[T1]{fontenc}    
\usepackage{hyperref}       
\usepackage{url}            
\usepackage{booktabs}       
\usepackage{amsfonts}       
\usepackage{nicefrac}       
\usepackage{microtype}      

\usepackage{amsmath}
\usepackage{dsfont} 
\usepackage{graphicx}
\usepackage{caption}
\usepackage{subcaption}
\usepackage{float} 
\usepackage{booktabs} 
\usepackage{multirow} 
\usepackage[table]{xcolor}
\usepackage{flushend}
\usepackage{algorithm}
\usepackage{algorithmic}
\usepackage{tikz}
\usetikzlibrary{bayesnet} 

\usepackage[round,numbers,sort]{natbib}

\def \Real{{\mathbb R}} 
\def \Natural{{\mathbb N}} 
\newcommand{\eValue}[1]{\mathbb{E}\left\{ #1 \right\}}

\newcommand{\ie}{i.e., }

\newcommand{\eg}{e.g., }

\newcommand{\N}[1]{\mathcal{N}\left( #1\right)}

\newcommand{\Cat}[1]{{\rm Cat}\left( #1\right)}

\newcommand{\NIG}[1]{{\rm NIG}\left( #1\right)}

\newcommand{\Beta}[1]{{\rm Beta}\left( #1\right)}

\newcommand{\U}[1]{\mathcal{U}\left( #1\right)}

\newcommand{\argmax}{\mathop{\mathrm{argmax}}}

\newcommand{\Var}{\mathbb{V}\mathrm{ar}}

\title{Bayesian bandits: balancing the exploration-exploitation tradeoff via double sampling}

\author{ I\~{n}igo Urteaga and Chris H.~Wiggins\\
	{\sf \{inigo.urteaga, chris.wiggins\}@columbia.edu} \\\\
	Department of	Applied Physics and Applied Mathematics\\
	Data Science Institute\\
	Columbia University\\
	New York City, NY 10027
}

\begin{document}
	\maketitle

\begin{abstract}
Reinforcement learning studies how to balance exploration and exploitation in real-world systems, optimizing interactions with the world while simultaneously learning how the world operates. One general class of algorithms for such learning is the multi-armed bandit setting. Randomized probability matching, based upon the Thompson sampling approach introduced in the 1930s, has recently been shown to perform well and to enjoy provable optimality properties. It permits generative, interpretable modeling in a Bayesian setting, where prior knowledge is incorporated, and the computed posteriors naturally capture the full state of knowledge. In this work, we harness the information contained in the Bayesian posterior and estimate its sufficient statistics via sampling. In several application domains, for example in health and medicine, each interaction with the world can be expensive and invasive, whereas drawing samples from the model is relatively inexpensive. Exploiting this viewpoint, we develop a double sampling technique driven by the uncertainty in the learning process: it favors exploitation when certain about the properties of each arm, exploring otherwise. The proposed algorithm does not make any distributional assumption and it is applicable to complex reward distributions, as long as Bayesian posterior updates are computable. Utilizing the estimated posterior sufficient statistics, double sampling autonomously balances the exploration-exploitation tradeoff to make better informed decisions. We empirically show its reduced cumulative regret when compared to state-of-the-art alternatives in representative bandit settings.
\end{abstract}

\section{Introduction}
\label{sec:introduction}

In a plethora of problems in science and engineering, one needs to decide which action to take next, based on partial information about the options available: a doctor must prescribe a medicine to a patient, a manager must allocate resources to competing projects, an ad serving algorithm must decide where to place ads, etc. In practice, the underlying properties of each choice are only partially known at the time of the decision, but one hopes that the understanding of the caveats involved will improve as time passes.

This set of problems has an illustrative gambling analogy, where a person facing a row of slot machines needs to devise its playing strategy (policy): which arms to play and in which order. The aim is to maximize the expected reward after a certain set of actions. Statisticians have studied this abstraction under the name of the multi-armed bandit problem for decades, e.g., in the seminal works by \citet{j-Robbins1952,j-Robbins1956}. The multi-armed bandit setting consists of sequential interactions with the world with rewards that are independent and identically distributed, or the related contextual bandit case, in which the reward distribution depends on different information or `context' presented with each interaction. It has played an important role in many fields across science and engineering.

Several algorithms have been proposed to overcome the exploration-exploitation tradeoff in such problems, mostly based on heuristics, on upper confidence bounds, or on the Gittins index. From the former, the $\epsilon$-greedy approach (randomly pick an arm with probability $\epsilon$, otherwise be greedy) has become very popular due to its simplicity, while nonetheless retaining often good performance \cite{j-Auer2002}. In the latter case, \citet{j-Gittins1979} formulated a method based on computing the optimal strategy for certain types of bandits, where geometrically discounted future rewards are considered. There are several difficulties inherent to the exact computation of the Gittins index and thus, approximations have been developed as well \cite{j-Brezzi2002}. These and other intrinsic challenges of the method have limited its applicability \cite{b-Sutton1998}.

\citet{j-Lai1985} introduced another class of algorithms, based on upper confidence intervals of the expected reward of each arm, for which strong theoretical guarantees were proved \cite{j-Lai1987}. Nevertheless, these algorithms might be far from optimal in the presence of dependent and more general reward distributions \cite{j-Scott2010}. Bayesian counterparts of UCB-type algorithms have been proposed in \cite{ip-Kaufmann2012}, where they show it provides an unifying framework for other variants of the UCB algorithm for distinctive bandit problems.

Recently, the problem has re-emerged both from a practical (importance in e-commerce and web applications, \eg \cite{j-Li2010}) and a theoretical (research on probability matching algorithms and their regret bounds, \eg \cite{j-Agrawal2011} and \cite{ip-Maillard2011}) point of view.

Contributing to this revival was the observation that
one of the oldest heuristics to address the exploration-exploitation tradeoff, \ie Thompson sampling \cite{j-Thompson1935,j-Thompson1933}, has been empirically proven to perform satisfactorily (see \cite{ic-Chapelle2011} and \cite{j-Scott2015} for details). Contemporaneously, theoretical study established several performance bounds, both for problems with and without context \cite{j-Agrawal2012,j-Agrawal2012a,ic-Korda2013,j-Russo2014,j-Russo2016}.

In this work, we are interested in the randomized probability matching approach, as it connects to the Bayesian learning paradigm. It readily facilitates not only generative and interpretable modeling, but sequential and batch processing algorithm development too.

Specifically, we investigate the benefits of fully harnessing the posteriors obtained via the Bayesian sequential learning process. We hereby avoid distributional assumptions to allow for complicated relationships among action rewards, as long as Bayesian posterior updates are computable.

We explore the benefits of sampling the model posterior to estimate the sufficient statistics that drive randomized probability matching algorithms. Our motivation is cases where sampling from the model posterior is inexpensive relative to interacting with the world, which may be expensive or invasive or, as in the medical application domain, both. The goal is that, with informative posterior sufficient statistics, better decisions can be made, leading to a lower cumulative regret.

We propose a double sampling technique for the multi-armed bandit problem, based on (1) Monte Carlo sampling, to approximate otherwise unsolvable integrals, and (2), a sampling-based arm-selection policy.

The policy is driven by the uncertainty in the learning process, as it favors exploitation when certain about the properties of the arms, exploring otherwise. Due to this autonomous exploration-exploitation balancing technique, the proposed algorithm achieves improved average performance, with important regret reductions.

We formally introduce the problem in Section \ref{sec:problem_formulation}, before providing all the details of our proposed double sampling method in Section \ref{sec:proposed_method}. The performance of double sampling is compared to the Thompson sampling and Bayes-UCB alternatives in Section \ref{sec:evaluation}, and we conclude with final remarks in Section \ref{sec:conclusion}.

\section{Problem formulation}
\label{sec:problem_formulation}

We mathematically formulate the multi-armed bandit problem as follows. Let $a\in\{1,\cdots,A\}$ indicate the arms of the bandit (possible actions to take), and $f_{a}(y|\theta)$ the stochastic reward distribution of each arm. For every time instant, the observed reward $y_t$ is independently drawn from the reward distribution corresponding to the played arm. We denote as $a_t$ the arm played at time instant $t$; $a_{1:t} \equiv (a_1, \cdots , a_t)$ refers to the sequence of arms played up to time $t$, and similarly, $y_{1:t} \equiv (y_1, \cdots , y_t)$ to the sequence of observed rewards.

In the multi-armed bandit setting one must decide, based on observed rewards $y_{1:t}$ and actions $a_{1:t}$, which arm to play next in order to maximize rewards. Due to the stochastic nature of the rewards, their expectation under the arm's distribution is the statistic of interest. We denote each arm's expected reward as $\mu_{a}(\theta)=\mathbb{E}_{a}\{y|\theta\}$, which is parameterized by the arm-dependent parameters $\theta$.

When the properties of the arms (\ie their parameters) are known, one can readily determine the optimal selection policy, \ie
	\begin{equation}
	a^*(\theta)=\argmax_{a}\mu_{a}(\theta) \; .
	\end{equation}
However, the optimal solution for the multi-armed bandit is only computable in closed form in very few special cases \cite{j-Bellm1956, j-Gittins1979}, and it fails to generalize to more realistic reward distributions and scenarios \cite{j-Scott2010}. The biggest challenge occurs when the parameters are unknown, as one might end up playing the wrong arm forever if incomplete learning occurs \cite{j-Brezzi2000}. 

Amongst the different algorithms to overcome these issues, the randomized probability matching, \ie playing each arm in proportion to its probability of being optimal, is a particularly appealing one. It has shown to be easy to implement, efficient and broadly applicable.

Given the parameters $\theta$, the expected reward of each arm is deterministic and, thus, one must pick the arm with the maximum expected reward
\begin{equation}
\mathrm{Pr}\left[a=a_{t+1}^*|a_{1:t}, y_{1:t}, \theta \right] = \mathrm{Pr}\left[a=a_{t+1}^*|\theta \right] = I_a(\theta),
\label{eq:theta_known_pr_arm_optimal}
\end{equation}
where we use the indicator function
\begin{equation}
I_a(\theta) = \begin{cases}
1, \; \mu_{a}(\theta)=\max\{\mu_1(\theta), \cdots, \mu_A(\theta)\} \;, \\
0, \; \text{otherwise} \;.
\end{cases}
\label{eq:indicator_arm_optimal}
\end{equation}

Under random probability matching, the aim is to compute the probability of a given arm $a$ being optimal for the next time instant, $p_{a,t+1}\in [0,1]$, even with unknown parameters. Mathematically,
\begin{equation}
\begin{split}
p_{a,t+1} &\equiv \mathrm{Pr}\left[a=a_{t+1}^* \big| a_{1:t}, y_{1:t}\right] \\
	&\equiv \mathrm{Pr}\left[ \mu_{a}(\theta) = \max\{\mu_1(\theta), \cdots, \mu_A(\theta)\} \big| a_{1:t}, y_{1:t}\right].
\end{split}
\label{eq:theta_unknown_pr_arm_optimal}
\end{equation}

Note that there is an inherent uncertainty about the unknown properties of the arms, as Eqn.~\eqref{eq:theta_unknown_pr_arm_optimal} is parameterized by $\theta$. In order to compute a solution to this problem, recasting it as a Bayesian learning problem, where $\theta$ is a random variable, is of great help. It allows for computation of posterior and marginal distributions, with direct connection to sampling techniques.

\section{Proposed method: double sampling}
\label{sec:proposed_method}

The multi-armed bandit problem consists of two separate but intertwined tasks: (1) learning about the properties of the arms, and (2) deciding what arm to play next. The problem is sequential in nature, as one makes a decision on which arm to play and learns from the observed reward, one observation at a time.

We cast the multi-armed bandit problem as a sequential Bayesian learning task. By doing so, we capture the full state of knowledge about the world at every time instant. We incorporate any available prior information to the learning process, and update our knowledge about the unknown parameter $\theta$, as we sequentially play arms and observe rewards. This learning can be done both sequentially or in batches, as Bayesian posterior updates are computable for both cases \cite{b-Bernardo2009}.

However, the solution to the probability matching equation in \eqref{eq:theta_unknown_pr_arm_optimal} is analytically intractable, so we approximate it via Monte Carlo sampling. For balancing the exploration-exploitation tradeoff, we propose a sampling-based probability matching technique too. The proposed arm-selection policy is a function of the uncertainty in the learning process. The intuition is that we exploit only when certain about the properties of the arms, while we keep exploring otherwise.

We elaborate on the foundations of the proposed double sampling method in the following sections, before presenting it formally in Algorithm \autoref{alg:bayesianDoubleSampling}.

\subsection{Bayesian multi-armed bandits}
\label{ssec:bayesian_multi_armed_bandit}

We are interested in computing, after playing arms $a_{1:t}$ and observing rewards $y_{1:t}$, the probability $p_{a,t+1}$ of each arm $a$ being optimal for the next time instant. In practice, one needs to account for the lack of knowledge of each arm's properties, \ie the unknown parameter $\theta$ in Eqn.~\eqref{eq:theta_unknown_pr_arm_optimal}.

We do so by following the Bayesian methodology, where the parameters are considered to be another set of random variables. The uncertainty over the parameters can be accounted for by marginalizing over their probability distribution.

Specifically, we marginalize over the posterior of the parameters after observing rewards and actions up to time $t$,
\begin{equation}
\begin{split}
p_{a,t+1} \equiv \mathrm{Pr}\left[a=a_{t+1}^* \big| a_{1:t}, y_{1:t}\right] &= f(a=a^*_{t+1}|a_{1:t}, y_{1:t}) \\
&=\int f(a=a^*_{t+1}|a_{1:t}, y_{1:t}, \theta) f(\theta|a_{1:t}, y_{1:t}) \mathrm{d}\theta \;.
\end{split}
\label{eq:pr_arm_optimal_bayes}
\end{equation}

Given a prior for the parameters $f(\theta)$ and the per-arm reward distribution $f_{a}(y|\theta)$, one can compute the posterior of each arm's parameters by
\begin{equation}
\begin{split}
f(\theta|a_{1:t}, y_{1:t}) &\propto f_{a_t}(y_t | \theta)f(\theta | a_{1:t-1}, y_{1:t-1}) \\
& \propto \left[\prod_{\tau=1}^t f_{a_{\tau}}(y_{\tau}|\theta)\right] f(\theta) \; .
\end{split}
\label{eq:seq_param_posterior}
\end{equation}

This posterior provides information (with uncertainty) about the characteristics of the arm. Note that the updates can usually be written in both sequential and batch forms. This flexibility is of great help in many practical scenarios, as one can learn from historic observations, as well as process data as it comes.

Even if analytical expressions for the parameter posteriors are available for many models of interest, computing the probability of any given arm being optimal is analytically intractable, due to the nonlinearities induced by the indicator function as in Eqn.~\eqref{eq:indicator_arm_optimal}
\begin{equation}
\begin{split}
p_{a,t+1} &=\int f(a=a^*_{t+1}|a_{1:t}, y_{1:t}, \theta) f(\theta|a_{1:t}, y_{1:t}) \mathrm{d}\theta = \int I_a(\theta) f(\theta|a_{1:t}, y_{1:t}) \mathrm{d}\theta \; .
\label{eq:pr_arm_optimal_bayes_indicator}
\end{split}
\end{equation}

\subsection{Monte-Carlo integration}
\label{ssec:mc_integration}

We harness the power of Monte Carlo sampling to compute the otherwise analytically intractable integral in Eqn.~\eqref{eq:pr_arm_optimal_bayes_indicator}. We obtain a Monte Carlo based random measure approximation to compute estimates of $p_{a,t+1}\in [0,1]$ as follows:
\begin{enumerate}
	\item Draw $M$ parameter samples from the updated posterior distribution
	\begin{equation}
	\theta^{(m)}\sim f(\theta|a_{1:t}, y_{1:t}), \; \; m=\{1, \cdots, M\} \; .
	\end{equation}
	\item For each parameter sample $\theta^{(m)}$, compute the expected reward and determine the best arm 
	\begin{equation}
	a_{t+1}^*(\theta^{(m)})=\argmax_{a}\mu_{a}(\theta^{(m)}) \; .
	\end{equation}
	\item Define the random measure approximation
	\begin{equation}
	f(a =a_{t+1}^*|a_{1:t}, y_{1:t}) \approx f_M(a =a_{t+1}^*|a_{1:t}, y_{1:t}) \approx \frac{1}{M} \sum_{m=1}^M \delta\left(a - a_{t+1}^*(\theta^{(m)}) \right) ,
	\label{eq:pr_arm_optimal_bayes_MC}
	\end{equation}
	where $\delta(\cdot)$ denotes the Dirac delta function.
	\item Estimate the first- and second-order sufficient statistics of $f_M(a =a_{t+1}^*|a_{1:t}, y_{1:t})$, \ie 
	\begin{equation}
	\begin{cases}
	\hat{p}_{a,t+1}=\eValue{f_M(a =a_{t+1}^*|a_{1:t}, y_{1:t})} =\frac{1}{M}\sum_{m=1}^M I_a\left(\theta^{(m)}\right) \; , \\
	\hat{\sigma}^2_{a,t+1}=\Var\{f_M(a =a_{t+1}^*|a_{1:t}, y_{1:t})\} =\frac{1}{M} \sum_{m=1}^M \left(I_a\left(\theta^{(m)}\right)- \hat{p}_{a,t+1} \right)^2 \; .
	\end{cases}
	\label{eq:pr_arm_optimal_bayes_MC_suff_statistics}
	\end{equation}
	\item Estimate which is the optimal arm and with what probability
	\begin{equation}
	\begin{cases}
	\hat{a}_{t+1}^* =\argmax_{a} \hat{p}_{a,t+1} \; ,  \\
	\hat{p}^*_{a,t+1}=\max_{a} \hat{p}_{a,t+1} \; .
	\end{cases}
	\end{equation}
\end{enumerate}

\subsection{Sampling-based policy}
\label{ssec:sampling_policy}

In any bandit setting, given the available information at time $t$, one needs to decide which arm to play next. A randomized probability matching technique would pick the next arm $a$ with probability $p_{a,t+1}$. On the contrary, a greedy approach would choose the arm with the highest probability of being optimal, \ie $p^*_{a,t+1}$.

We present an alternative sampling-based probability matching arm-selection policy that finds a balance between these two cases. We rely on the Monte Carlo approximation to Eqn.~\eqref{eq:pr_arm_optimal_bayes_indicator}, and leverage the estimated sufficient statistics in Eqn.~\eqref{eq:pr_arm_optimal_bayes_MC_suff_statistics} to balance the exploration-exploitation tradeoff. We draw candidate arm samples from the random measure in Eqn.~\eqref{eq:pr_arm_optimal_bayes_MC}, and automatically adjust the probability matching technique according to the accuracy of this approximation.

The number of candidate arm samples drawn is instrumental for our sampling-based policy. We automatically adjust its value according to the uncertainty on the optimality of each arm, \ie $\sigma^2_{a,t+1}$. By doing so, we account for the uncertainty of the learning process in the arm-selection policy, dynamically balancing exploration and exploitation.

The number of candidate arm samples to draw is inversely proportional to the probability of not picking the optimal arm. We denote this probability as $p_{FA}$, which is computed for each arm as
\begin{equation}
p_{FA}^{(a)} =Pr\left(p_{a,t+1} > p^*_{a,t+1} \right) = 1 - F_{p_{a,t+1}}(p^*_{a,{t+1}}) \; ,
\end{equation}
where $p_{a,t+1}^*=\max_{a}p_{a,t+1}$. The true cumulative density function $F_{p_{a,t+1}}(\cdot)$ is analytically intractable as well, but we approximate it (based on the central limit theorem guarantees of the MC estimates) with a Gaussian truncated to the CDF's range
\begin{equation}
F_{p_{a,t+1}}(p^*_{a,{t+1}}) \approx \Phi_{[0,1]}\left(\frac{p^*_{a,{t+1}}-p_{a,t+1}}{\sigma_{a,t+1}}\right) \;.
\end{equation}
Since we can not exactly evaluate $p_{a,t+1}$ and $\sigma_{a,t+1}$, we resort to our Monte Carlo estimates in Eqn.~\eqref{eq:pr_arm_optimal_bayes_MC_suff_statistics} instead. 

All in all, the proposed sampling policy proceeds as follows:
\begin{enumerate}
	\item Determine $N_{t+1}$, the number of candidate arm samples to draw
		\begin{equation}
		\begin{split}
		N_{t+1} \propto & \log\left(\frac{1}{p_{FA}}\right), \; \; p_{FA}=\frac{1}{K-1}\sum_{a \neq \hat{a}_{t+1}^*}p_{FA}^{(a)}\;, \\
		& p_{FA}^{(a)} \approx 1- \Phi_{[0,1]}\left(\frac{\hat{p}^*_{a,{t+1}}-\hat{p}_{a,t+1}}{\hat{\sigma}_{a,t+1}}\right) \; .
		\end{split}
		\label{eq:policy_n_samples}
		\end{equation}
	\item Draw $N_{t+1}$ candidate arm samples
	\begin{equation}
	\hat{a}_{t+1}^{(n)} \sim \Cat{\hat{p}_{a,t+1}}, \; \; \;  n=1,\cdots, N_{t+1} \; .
	\end{equation}
	\item Pick the most probable optimal arm, given drawn candidate arm samples $\hat{a}_{t+1}^{(n)}$
	\begin{equation}
	a_{t+1} = \text{Mode}\left(\hat{a}_{t+1}^{(n)}\right), \; \; \; n=1,\cdots,N_{t+1} \;.
	\end{equation}
\end{enumerate}

By allowing for $N_{t+1}$ to be adjusted based upon the uncertainty of the learning process, we balance the exploration-exploitation tradeoff. We present full details of the proposed double sampling technique in Algorithm \autoref{alg:bayesianDoubleSampling}.

\begin{algorithm}
	\begin{algorithmic}[1]
		\REQUIRE Number of arms $A$, number of MC samples $M$, and horizon $T$
		\REQUIRE Prior over model parameters $f(\theta)$ and per-arm reward distributions $f_a(y|x,\theta)$
		\STATE $D=\emptyset$
		\FOR{$t=1, \cdots, T$}
		\STATE Draw $M$ posterior parameter samples
		\begin{equation}
		\theta_{t+1}^{(m)}\sim f(\theta|a_{1:t}, x_{1:t}, y_{1:t}), \; \; \; m=\{1, \cdots, M\}
		\end{equation}
		\STATE If applicable, receive context $x_{t+1}$
		\FOR{$a=1, \cdots, A$}
		\STATE Compute expected reward, \\
		\qquad \qquad per parameter sample
		\begin{equation}
		\mu_{a,t+1}(\theta_{t+1}^{(m)})=\mu_{a}(x_{t+1},\theta_{t+1}^{(m)})
		\end{equation}
		\STATE Compute sufficient statistics
		\begin{equation}
		\begin{split}
		&\qquad \; \; \hat{p}_{a,t+1}=\frac{1}{M}\sum_{m=1}^M I_a\left(\theta_{t+1}^{(m)}\right) \\
		&\qquad \; \; \hat{\sigma}^2_{a,t+1}=\frac{1}{M} \sum_{m=1}^M \left(I_a\left(\theta_{t+1}^{(m)}\right)- \hat{p}_{a,t+1} \right)^2
		\end{split}
		\end{equation}
		\ENDFOR
		\STATE Compute estimates
		\begin{equation}
		\begin{cases}
		\hat{a}_{t+1}^* =\argmax_{a} \hat{p}_{a,t+1} \; ,  \\
		\hat{p}^*_{a,t+1}=\max_{a} \hat{p}_{a,t+1} \; .
		\end{cases}
		\end{equation}
		\FOR{$a=1,\cdots, A$}
		\STATE Compute probability of arm not being optimal
		\begin{equation}
		\hat{p}_{FA}^{(a)}=Pr\left(\hat{p}_{a,t+1} > \hat{p}^*_{a,t+1} \right)
		\end{equation}
		\ENDFOR
		\STATE Compute the number of candidate arm samples
		\begin{equation}
		N_{t+1} \propto \log\left(\frac{1}{\hat{p}_{FA}}\right), \; \; \; \hat{p}_{FA}=\frac{1}{A-1}\sum_{a \neq \hat{a}_{t+1}^*}\hat{p}_{FA}^{(a)}
		\end{equation}
		\STATE Draw $N_{t+1}$ candidate arm samples
		\begin{equation}
		\hat{a}_{t+1}^{(n)} \sim \Cat{\hat{p}_{a,t+1}}, \; \; \; n=1,\cdots, N_{t+1} 
		\end{equation}
		\STATE Play arm
		\begin{equation}
		a_{t+1} = \text{Mode}\left(\hat{a}_{t+1}^{(n)}\right), \; \; \; n=1,\cdots,N_{t+1}
		\end{equation}
		\STATE Observe reward $y_{t+1}$
		\STATE Update $D=D \cup \left\{x_{t+1}, a_{t+1}, y_{t+1}\right\}$
		\ENDFOR
	\end{algorithmic}
	\caption{Double sampling algorithm}
	\label{alg:bayesianDoubleSampling}
\end{algorithm}

The proposed sampling policy reduces to a probabilistic matching regime when uncertain about the arms, (\ie $N_t \approx 1$), but favors exploitation ($N_t \gg 1$) when the probability of picking a suboptimal arm is low. In other words, double sampling exploits only when confident about the learned probabilities ($\hat{\sigma}_{a,t+1} \rightarrow 0, N_t \gg 1)$, and picks the arm with the highest probability $\hat{p}_{a,t+1}$. However, for $N_t \approx 1$, a randomized probability matching is in play. Note that Thompson sampling is a special case of double sampling, when $N_{t}=1 \;, \forall t$.

To conclude, note that the sampling-based policy decides on the action to take next, by drawing from an approximation to the posterior density $p_{a,t+1}$. Precisely, by probability matching the expected return of each arm, which is estimated via Monte Carlo as in Eqn.~\eqref{eq:pr_arm_optimal_bayes_MC_suff_statistics}. For the derivation of performance bounds in multi-armed bandit problems and, in particular, regret bounds for posterior sampling techniques, one studies the expected returns of the arms. Due to Monte Carlo guarantees on the convergence of the computed estimates ($\lim_{M \to \infty} \hat{p}_{a,t+1} = p_{a,t+1}$), and the random probability matching nature of double sampling, the regret bounds for our proposed technique are of the same order as those of any posterior sampling technique \cite{j-Agrawal2012a,j-Russo2016}. We argue that the discrepancies are on the multiplicative constants, which we evaluate in the following section.

\section{Evaluation}
\label{sec:evaluation}

We now empirically evaluate the performance of double sampling in both discrete and continuous contextual multi-armed bandit settings. We compare the performance of our proposed algorithm, to that of Thompson sampling \cite{ic-Chapelle2011} and Bayes-UCB \cite{ip-Kaufmann2012}.

On the one hand, \cite{ic-Chapelle2011} show empirically the significant advantages Thompson sampling offers for the Bernoulli and other cases, while theoretical guarantees are provided in \cite{j-Agrawal2012,j-Agrawal2012a,ic-Korda2013,j-Russo2014,j-Russo2016}. On the other, \cite{ip-Kaufmann2012} prove the asymptotic optimality of Bayes-UCB's finite-time regret bound for the Bernoulli case, and argue that it provides an unifying framework for several variants of the UCB algorithm for different bandit problems: parametric multi-armed bandits and linear Gaussian bandits.

We compare double sampling as in Algorithm \ref{alg:bayesianDoubleSampling} to these two state-of-the-art algorithms, in order to provide empirical evidence of the reduced cumulative regret of our proposed approach. We define cumulative regret as
\begin{equation}
R_t=\sum_{\tau=0}^t \eValue{\left(y^*_{\tau}-y_{\tau} \right)} = \sum_{\tau=0}^t \mu_\tau^*-\bar{y}_{\tau} \; ,
\end{equation}
where for each time instant $t$, $\mu_t^*$ denotes the expected reward of the optimal arm and $\bar{y}_{t}$ the empirical mean of the observed rewards under the executed policy. Note that even if the bandits considered are stationary (\ie parameters are not dynamic), the expected rewards are indexed with time to accommodate their dependency with potentially time-dependent contexts $x_t$.

\subsection{Bernoulli bandits}
\label{ssec:bernoulli_bandits}

Bernoulli bandits are well suited for applications with binary rewards (\ie success or failure of an action). The rewards of each arm are modeled as independent draws from a Bernoulli distribution with success probabilities $\theta_a$, \ie
\begin{equation}
f_a(y|\theta)=\theta_a^{y}(1-\theta_a)^{(1-y)} \; .
\end{equation}
For a Bernoulli reward distribution, the posterior parameter update can be computed using the conjugate prior distribution $f(\theta_a|\alpha_{a,0}, \beta_{a,0})=\Beta{\theta_a|\alpha_{a,0}, \beta_{a,0}}$. After observing actions $a_{1:t}$ and rewards $y_{1:t}$, the posterior parameter distribution follows an updated Beta distribution

\begin{equation}
f(\theta_a|a_{1:t}, y_{1:t}, \alpha_{a,0}, \beta_{a,0}) =f(\theta_a|\alpha_{a,t}, \beta_{a,t}) =\Beta{\theta_a|\alpha_{a,t}, \beta_{a,t}}  \; ,
\end{equation}
with sequential updates
\begin{equation}
\begin{cases}
\alpha_{a,t}=\alpha_{a,t-1} + y_{t} \cdot \mathds{1}[a_t=a] \; ,\\
\beta_{a,t}=\beta_{a,t-1} + (1 - y_{t}) \cdot \mathds{1}[a_t=a] \; ,
\end{cases} 
\end{equation}
or, alternatively, batch updates of the following form
\begin{equation}
\begin{cases}
\alpha_{a,t}=\alpha_{a,0} + \sum_{t|a_t=a} y_{t} \; ,\\
\beta_{a,t}=\beta_{a,0} + \sum_{t|a_t=a} (1-y_{t}) \; .
\end{cases}
\end{equation}

\begin{figure}[!h]
	\centering
	\begin{subfigure}[b]{0.49\textwidth}
		\includegraphics[width=\textwidth]{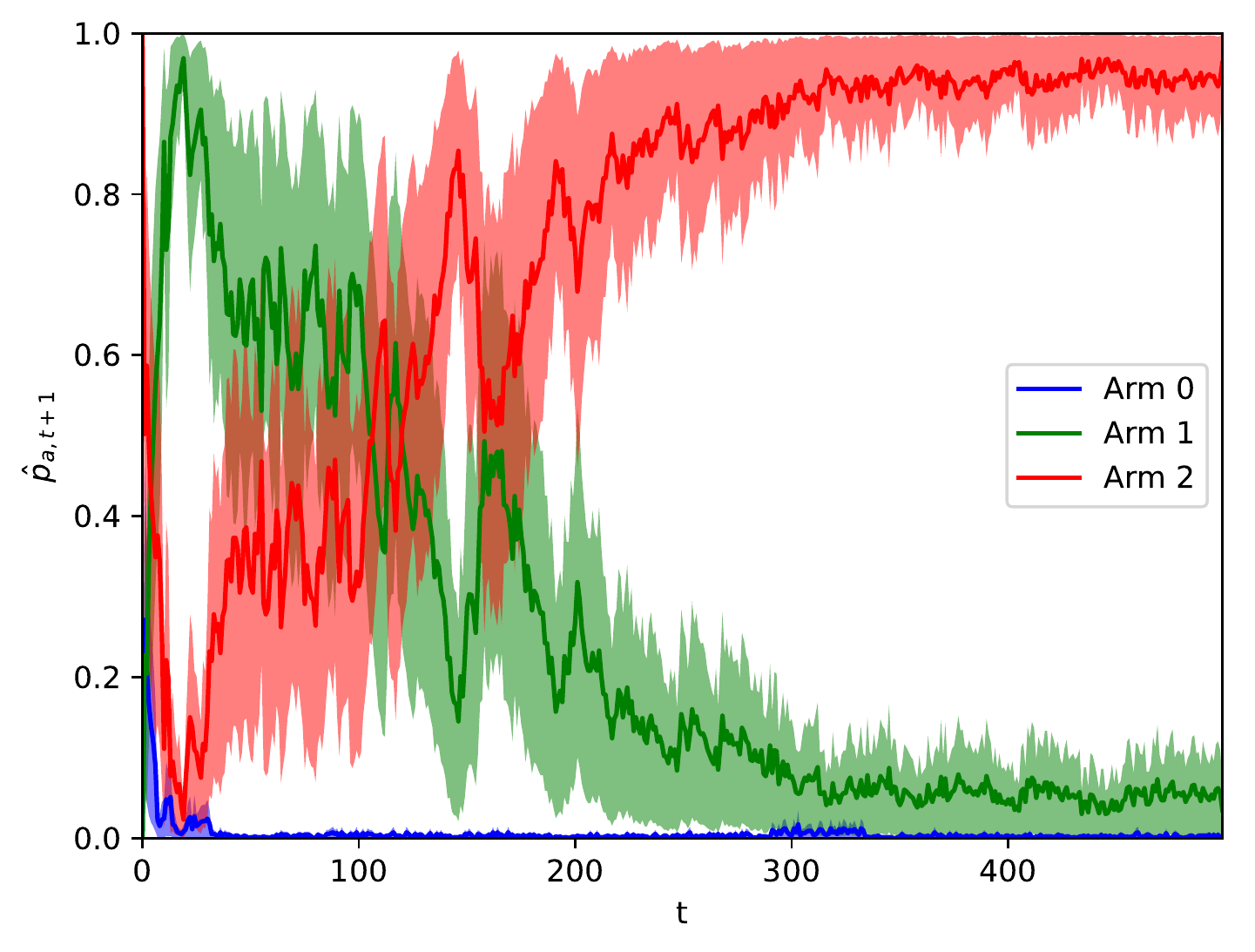}
		\caption{$\hat{p}_{a,t+1}$ over time.}
		\label{fig:pred_action_density}
	\end{subfigure}%
	\begin{subfigure}[b]{0.49\textwidth}
		\includegraphics[width=\textwidth]{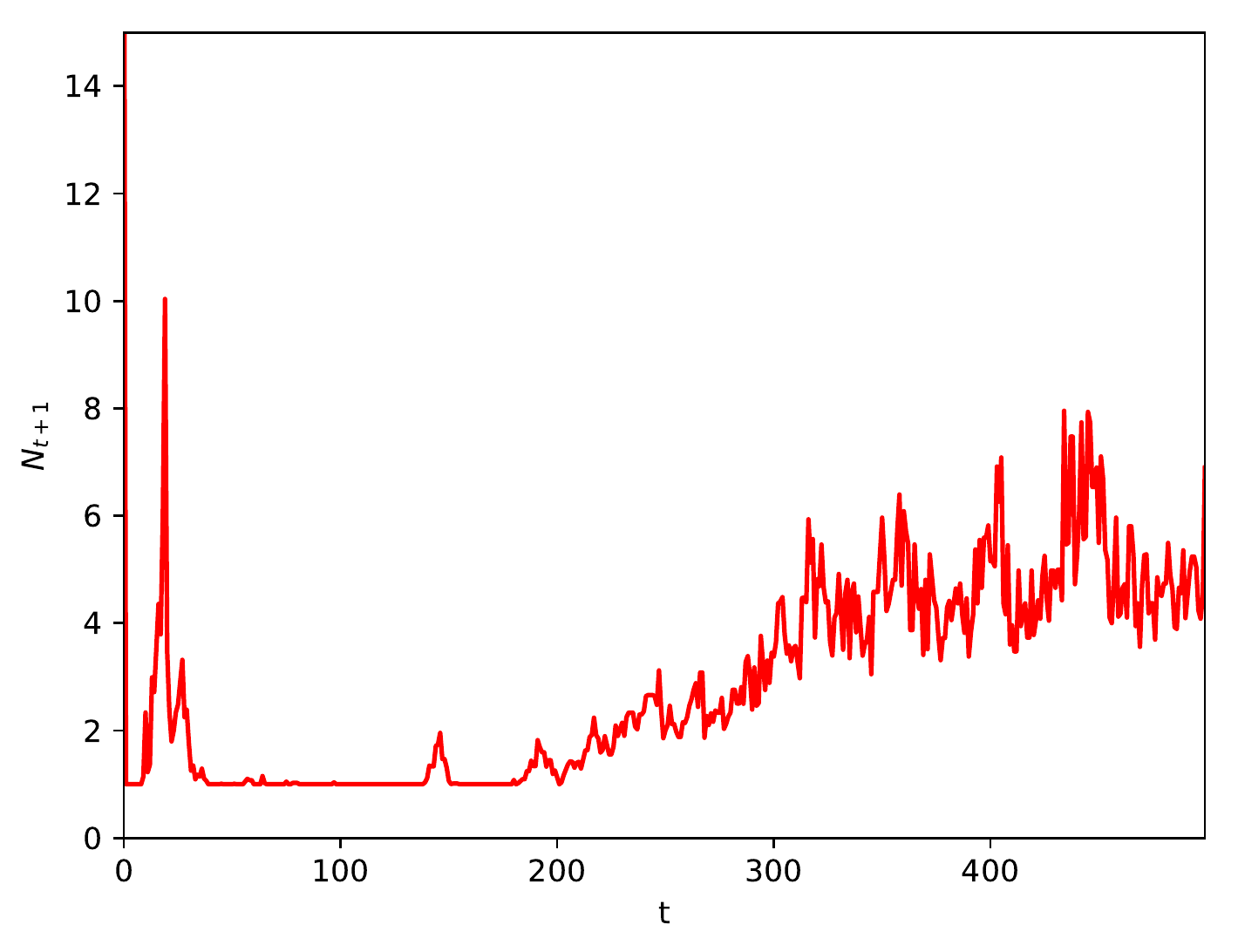}
		\caption{$N_{t+1}$ over time.}
		\label{fig:n_samples}
	\end{subfigure}
	\caption{Illustrative execution of double sampling.}
	\label{fig:approach_intuition}
\end{figure}

The sequential Bayesian learning process for a three-armed Bernoulli bandit with parameters $\theta=\left(0.4 \; 0.7 \; 0.8 \right)$ is illustrated in Fig. \ref{fig:pred_action_density}. We show the evolution of the probability of each arm being optimal as computed by our proposed algorithm: \ie the Monte Carlo approximation to $p_{a,t+1}$. For all results to follow, we use $M=1000$ Monte Carlo samples, as larger $M$s do not significantly improve regret performance. In Fig. \ref{fig:n_samples}, we illustrate how double sampling is {\em automatically} adjusted according to the uncertainty of the learning process, via the number of arm samples to draw (\ie $N_{t+1}$ in Eqn.~\eqref{eq:policy_n_samples}).

Let us elaborate on the exploration-exploitation tradeoff by following the double sampling bandit execution shown in Fig. \ref{fig:approach_intuition}. Observe how arm 0 is quickly discarded as a good candidate, while the decision over which of the other two arms is optimal requires further learning. For some time ($t<200$), there is high uncertainty about the properties of these two arms (high variance in Fig. \ref{fig:pred_action_density}). Thus, double sampling favors exploration ($N_{t+1}\approx 1$ in Fig. \ref{fig:n_samples}), until the uncertainty about which arm is best is reduced. Once the algorithm becomes more certain about the better reward properties of arm 2 ($t>200$), double sampling gradually favors a greedier policy ($N_{t+1}>1$).

All in all, within periods of high uncertainty, the number of samples $N_{t+1}$ is kept low (\ie exploration); on the contrary, when the learning is more accurate, it increases (\ie exploitation). By means of the double sampling technique, we account for the uncertainty in the learning process and thus, the proposed algorithm can reduce the variance over the actions taken.

\begin{figure}[!h]
	\centering
	\begin{subfigure}[b]{0.49\textwidth}
		\includegraphics[width=\textwidth]{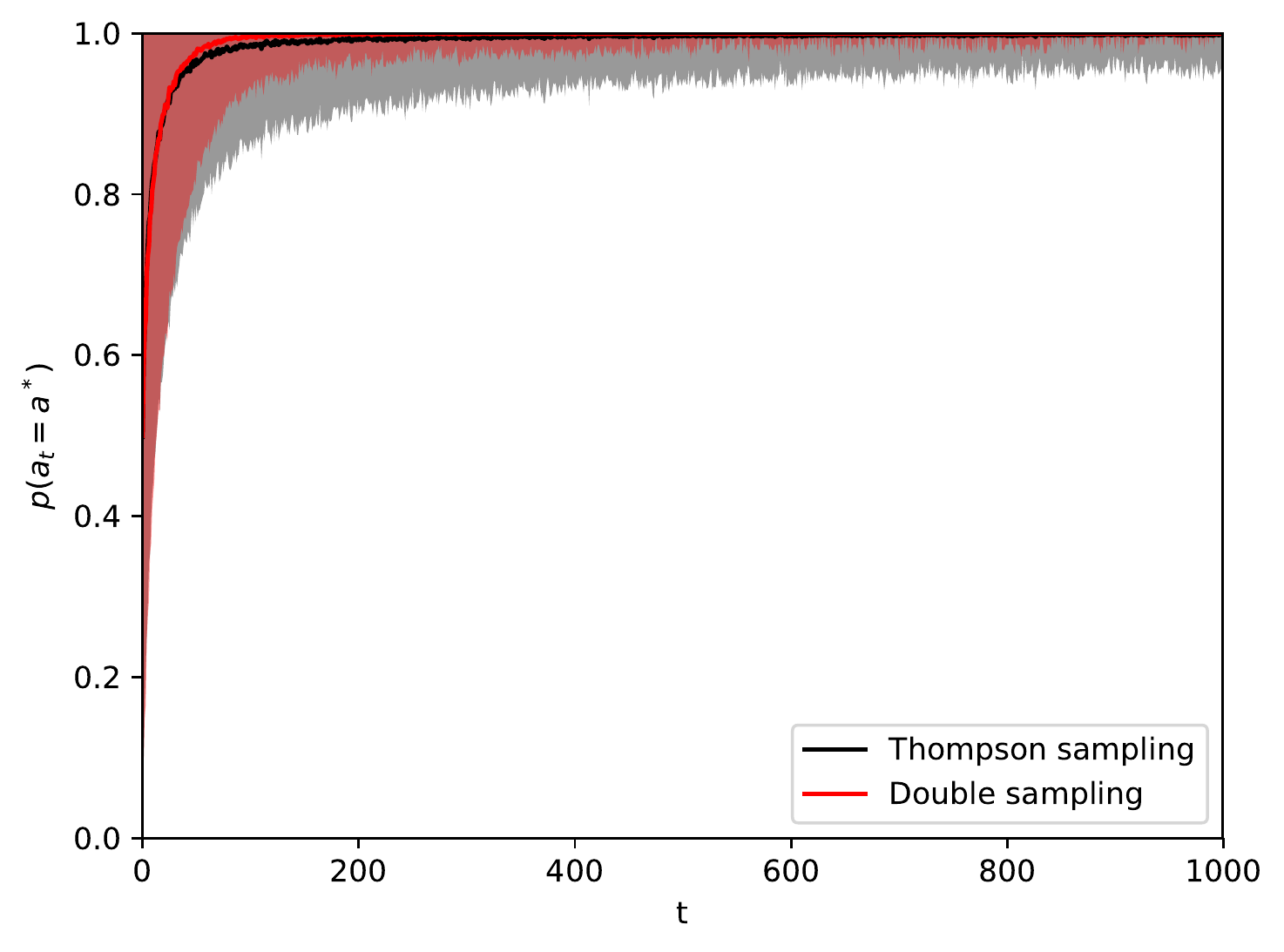}
		\label{fig:bernoulli_correct_actions_TS_DS}
	\end{subfigure}%
	\begin{subfigure}[b]{0.49\textwidth}
		\includegraphics[width=\textwidth]{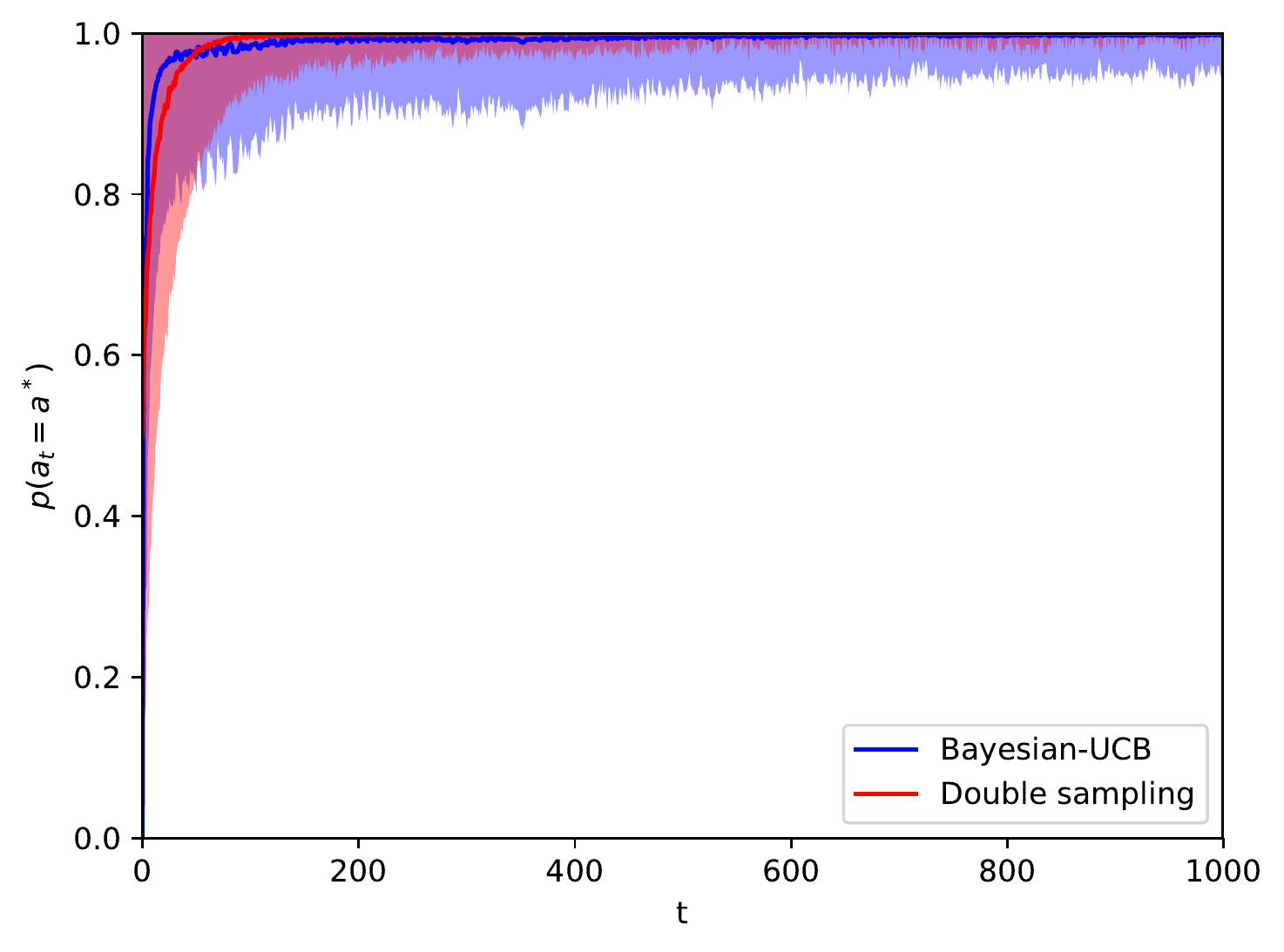}
		\label{fig:bernoulli_correct_actions_BUCB_DS}
	\end{subfigure}
	\caption{Averaged correct action probability (standard deviation as shaded region) with $A=2, \theta=\left(0.4 \; 0.8\right)$.}
	\label{fig:bernoulli_correct_actions_compare}
\end{figure}

We plot in Fig. \ref{fig:bernoulli_correct_actions_compare} the empirical probabilities of each algorithm playing the optimal arm over 5000 realizations\footnote{All averaged results in this work are computed over 5000 realizations of the same set of parameters.} of a Bernoulli bandit with $A=2, \theta=\left(0.4 \; 0.8\right)$. Observe how, even if in expectation all algorithms take the same actions, the action variability of double sampling is considerably smaller.

\begin{figure}[!h]
	\centering
	\includegraphics[width=0.55\textwidth]{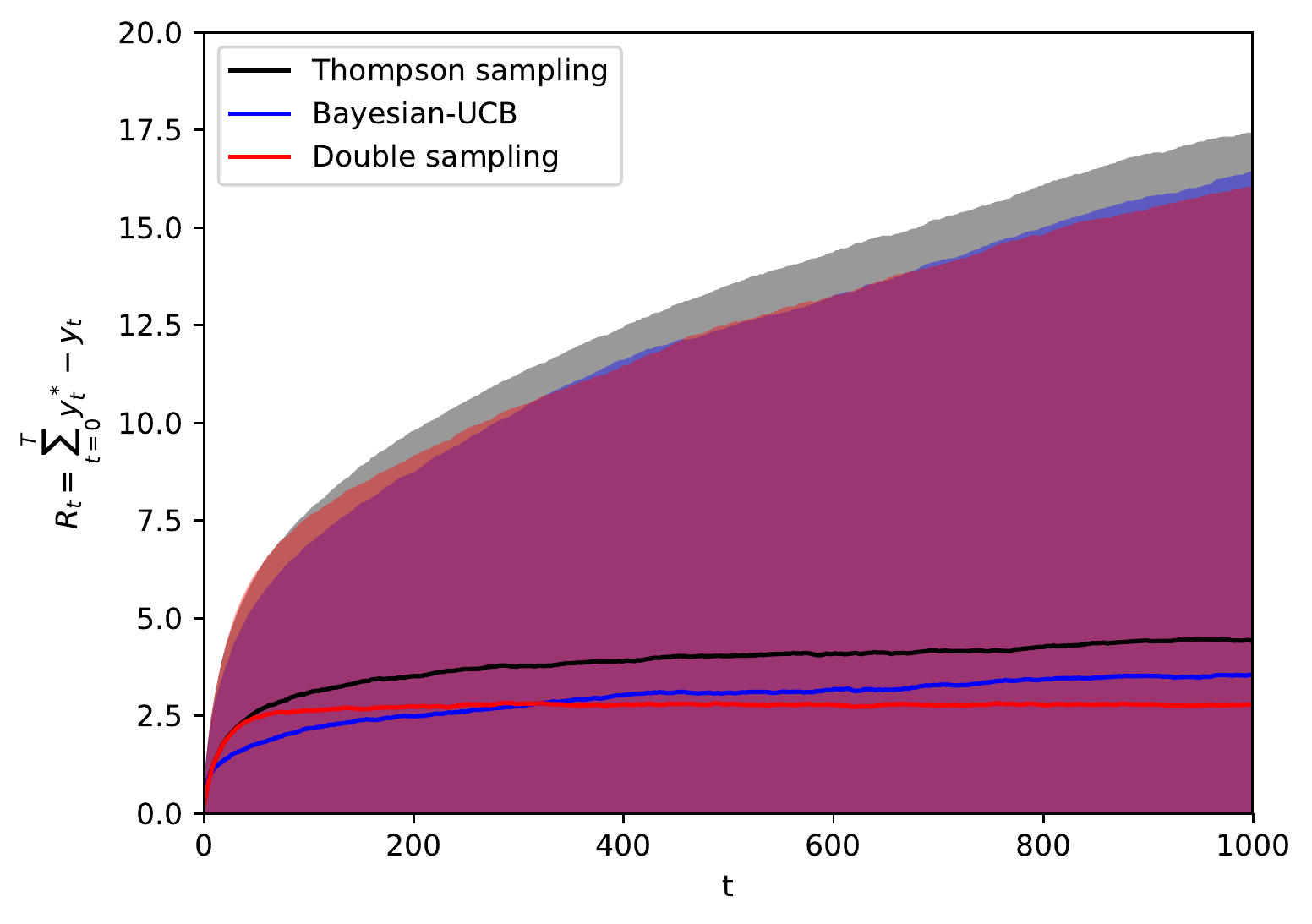}
	\caption{Averaged cumulative regret (standard deviation shown as shaded region) with $A=2, \theta=\left(0.4 \; 0.8\right)$.}
	\label{fig:bernoulli_cumulative_regret}
\end{figure}

As a result, the cumulative regret of our proposed technique is lower than those of the compared alternatives, \ie Thompson sampling and Bayes-UCB, (see averaged cumulative regrets in Fig. \ref{fig:bernoulli_cumulative_regret}). 

For any bandit problem, the difficulty of learning the properties of each arm is a key factor on its regret performance. Intuitively, this difficulty relates to how close the expected rewards of each arm are. Mathematically, it can be captured by the divergence between arm reward distributions. By computing the minimum Kullback-Leibler (KL) divergence between arms, one quantifies how ``difficult'' a multi-armed bandit problem is, as established by the regret lower-bound in \cite{j-Lai1985}. 

We evaluate the relative difference between the averaged cumulative regret of our proposed double sampling technique and the alternatives, \ie
\begin{equation}
\Delta_t^{(TS)} =\frac{R_{t}^{(DS)}}{R_{t}^{(TS)}}-1 \; \text{ and } \; \Delta_t^{(B-UCB)} =\frac{R_{t}^{(DS)}}{R_{t}^{(B-UCB)}}-1 \; ,
\label{eq:relative_cum_reg_dif}
\end{equation}
where $R_t^{(DS)}$ denotes the regret of the proposed double sampling approach at time $t$, $R_t^{(TS)}$, that of Thompson sampling, and $R_t^{(B-UCB)}$, that of Bayes-UCB.

\begin{figure}[!h]
	\centering
	\includegraphics[width=0.7\textwidth]{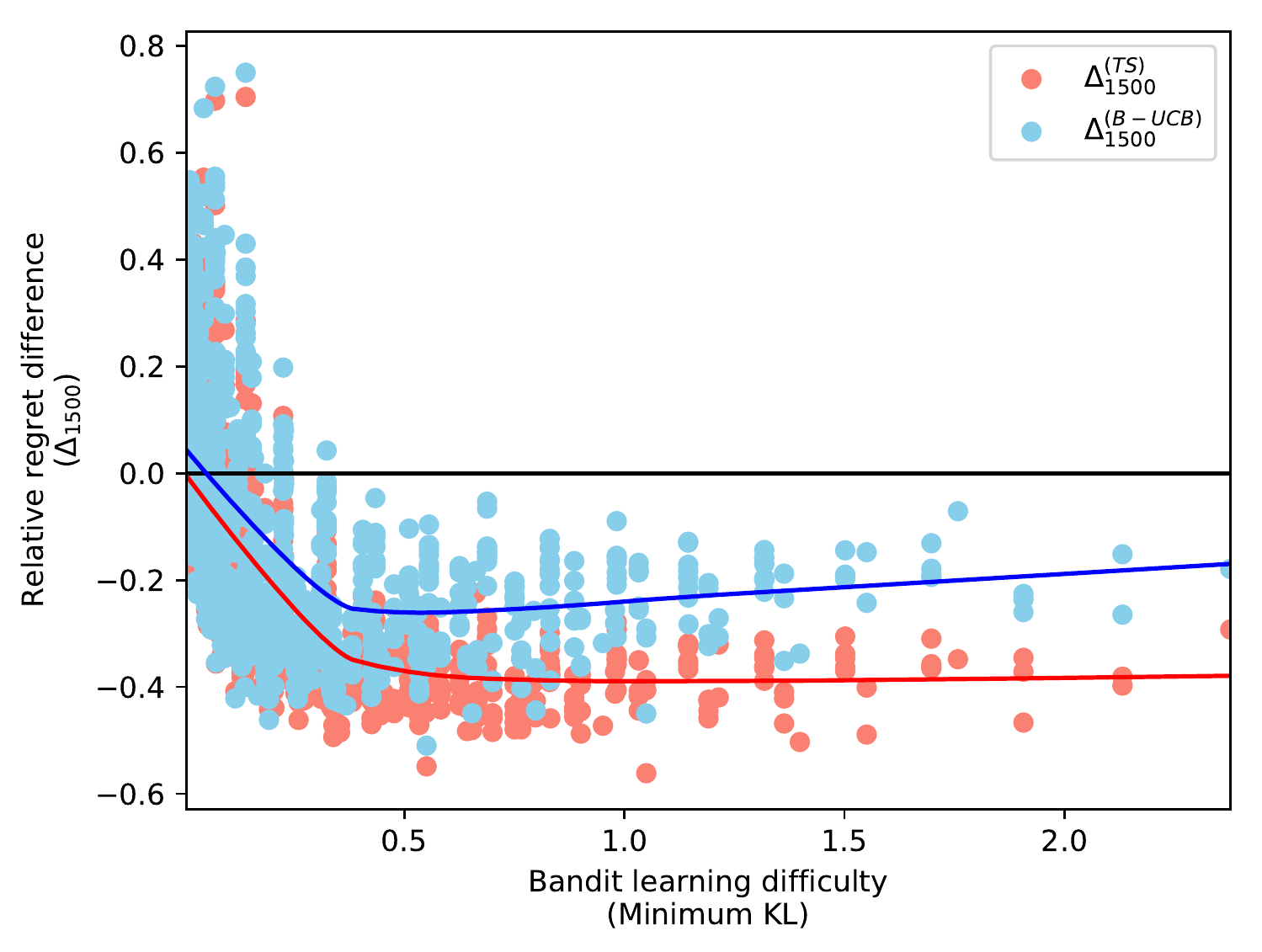}
	\caption{Relative average cumulative regret differences at $t=1500$.}
	\label{fig:bernoulli_relative_cumregret_kl}
\end{figure}

We show in Fig. \ref{fig:bernoulli_relative_cumregret_kl} results for the above metric indexed by the KL divergence of a wide range of Bernoulli bandit parameterizations\footnote{Bernoulli bandits with $A=2$ and $A=3$ arms, for all per-arm parameter permutations in the range $\theta_a\in[0,1]$ with grid size $0.05$.}. Note that the KL metric may map many parameter combinations to the same point in Fig. \ref{fig:bernoulli_relative_cumregret_kl}.

Double sampling performs significantly better than the alternatives when it is certain about the learned arm parameters. We obtain cumulative regret reductions around 25\% and 40\% when compared to B-UCB (with optimal quantile parameter $\alpha_t=1/t$ as in \cite{ip-Kaufmann2012}) and Thompson sampling, respectively. However, when the best arms are very similar ($KL<0.25$), performance worsens. First, recall that the regret lower-bound increases for all bandits with small KL values \cite{j-Lai1985}. Second, note that when the properties of the arms are very similar to each other, our algorithm resorts to a Thompson sampling-like policy ($N_{t+1}\approx1$), yielding near-equivalent performance.

Finally, we observe that for the challenging cases (\ie $KL<0.25$) cumulative regret shows high variance for the three alternatives (points are scattered in Fig. \ref{fig:bernoulli_relative_cumregret_kl}). On the contrary, when the difference between arm properties is distinguishable ($KL>0.25$), the proposed double sampling technique considerably reduces cumulative regret for Bernoulli multi-armed bandits.

\subsection{Contextual linear Gaussian bandits}
\label{ssec:contextLinearGaussian_bandits}

Another set of well studied bandits are those with continuous reward functions and, in particular, those with contextual dependencies. That is, the reward distribution of each arm is dependent on a time-varying $d$-dimensional context vector $x_t\in\Real^{d}$.

The contextual linear Gaussian bandit model is suited for these scenarios, where the expected reward of each arm is linearly dependent on the context $x\in\Real^{d}$, and the idiosyncratic parameters of the bandit $\theta\equiv\{w, \sigma\}$. That is, the per-arm reward distribution follows
\begin{equation}
f_a(y|x,\theta)=\N{y|x^\top w_a, \sigma_a^2} \; .
\end{equation}

For such reward distribution, the posterior can be computed with the Normal Inverse Gamma conjugate prior distribution
\begin{equation}
\begin{split}
f(w_a, \sigma_a^2|u_{a,0}, V_{a,0},\alpha_{a,0}, \beta_{a,0}) &= \NIG{w_a, \sigma_a^2|u_{a,0}, V_{a,0},\alpha_{a,0}, \beta_{a,0}} \\
& = \N{w_a|u_{a,0}, \sigma_a^2 V_{a,0}} \cdot \Gamma^{-1}\left(\sigma_a^2|\alpha_{a,0}, \beta_{a,0}\right) \;. \\
\end{split}
\end{equation}
Given previous actions $a_{1:t}$, contexts $x_{1:t}$ and rewards $y_{1:t}$, one obtains the following posterior 
\begin{equation}
f(w_a, \sigma_a^2|a_{1:t},y_{1:t},u_{a,0}, V_{a,0},\alpha_{a,0}, \beta_{a,0}) =\NIG{w_a, \sigma_a^2|u_{a,t}, V_{a,t},\alpha_{a,t}, \beta_{a,t}} \; ,
\end{equation}
where the parameters of the posterior are sequentially updated as
\begin{equation}
\begin{cases}
V_{a,t}^{-1} = V_{a,t-1}^{-1} + x_t x_t^\top \cdot \mathds{1}[a_t=a] \; ,\\
u_{a,t}= V_{a,t} \left( V_{a,t-1}^{-1} u_{a,t-1} + x_t y_{t}\cdot \mathds{1}[a_t=a]\right) \; ,\\
\alpha_{a,t}=\alpha_{a,t-1} + \frac{\mathds{1}[a_t=a]}{2} \; ,\\
\beta_{a,t}=\beta_{a,t-1} + \frac{\mathds{1}[a_t=a](y_{t_a}-x_t^\top\theta_{a,t-1})^2}{2\left(1+x_t^\top \Sigma_{a,t-1} x_t\right)} \; .
\end{cases}
\end{equation}
Alternatively, if data is collected in batches, one updates the posterior with
\begin{equation}
\begin{cases}
V_{a,t}^{-1}= V_{a,0}^{-1}+x_{{1:t}|t_a} x_{{1:t}|t_a}^\top \; ,\\
u_{a,t}=V_{a,t}\left(V_{a,0}^{-1}u_{a,0}+x_{{1:t}|t_a} y_{{1:t}|t_a}\right) \; ,\\
\alpha_{a,t}=\alpha_{a,0} + \frac{|t_a|}{2} \; ,\\
\beta_{a,t}=\beta_{a,0} + \frac{\left(y_{{1:t}|t_a}^\top y_{{1:t}|t_a} + u_{a,0}^\top V_{a,0}^{-1}u_{a,0} - u_{a,t}^\top V_{a,t}^{-1}u_{a,t} \right)}{2} \; ,
\end{cases}
\end{equation}
where $t_a=\{t|a_t=a\}$ indicates the set of time instances when arm $a$ is played.

We evaluate double sampling for the multi-armed contextual Gaussian bandit with uniform and uncorrelated context, \ie $x_{i,t}\sim \U{0,1}, i \in \{1, \cdots, d\}, t \in \Natural$. 

We again use the minimum KL divergence as a proxy for bandit complexity. The divergence is model agnostic, as many parameter combinations for any model may map to the same KL divergence value.

\begin{figure}[!h]
	\centering
	\includegraphics[width=0.5\textwidth]{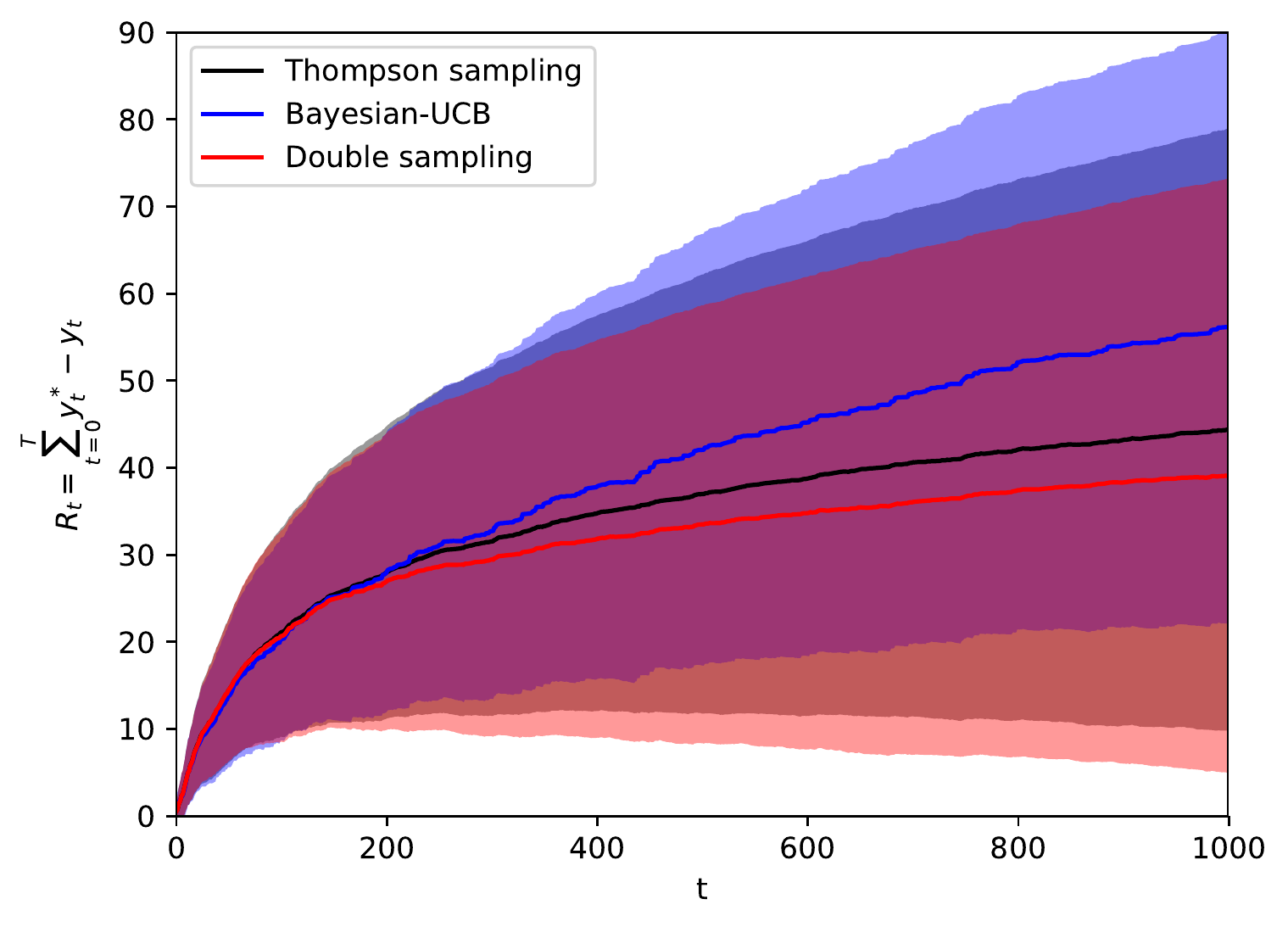}
	\caption{Averaged cumulative regret comparison (standard deviation shown as shaded region) with $A=2$, $w_0=(0.4 \; 0.4)^\top$, $w_1=(0.8 \; 0.8)^\top$, $\sigma_0=\sigma_1=0.2$.}
	\label{fig:linearGaussian_cumulative_regret_compare}
\end{figure}

\begin{figure}[!h]
	\centering
	\includegraphics[width=0.75\textwidth]{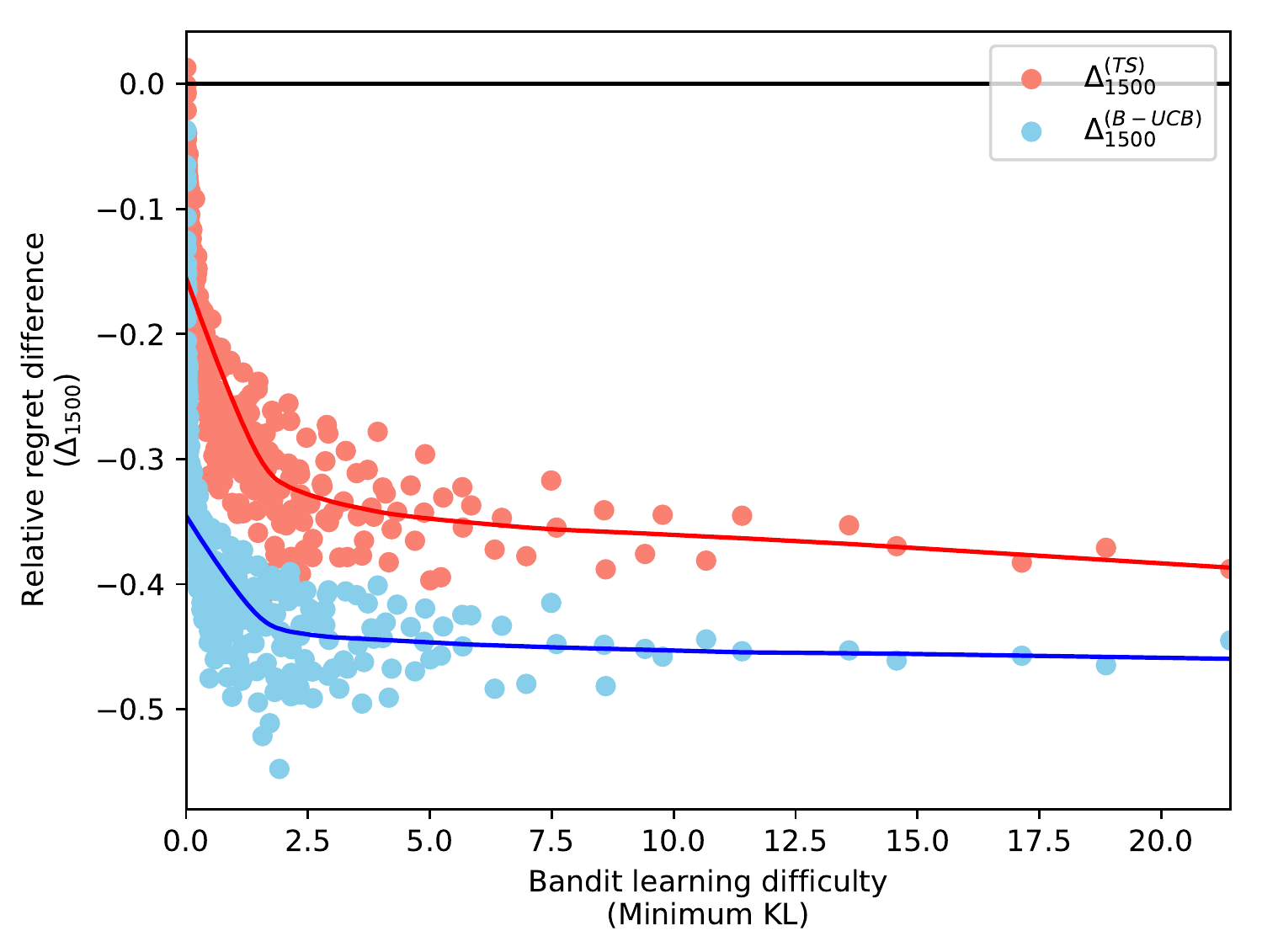}
	\caption{Relative average cumulative regret differences at $t=1500$.}
	\label{fig:linearGaussian_relative_cumregret_kl}
\end{figure}

We provide results for a specific two-armed contextual Gaussian bandit in Fig. \ref{fig:linearGaussian_cumulative_regret_compare}, and in Fig. \ref{fig:linearGaussian_relative_cumregret_kl}, average cumulative regret relative differences (as in Eqn.~\eqref{eq:relative_cum_reg_dif}) for a wide range of parameterizations\footnote{Contextual linear Gaussian bandits with per-dimension parameter $w_i \in [-1,1], i\in\{1,2\}$ with gaps of $0.1$, and $\sigma \in [0.1,1]$ with step size of $0.1$.} of two-dimensional contextual linear Gaussian bandits with two arms.

Again, when the reward difference between arms is easy to learn (KL>0.25), double sampling attains significant cumulative regret reductions. The regret improvement is most evident for models with significant KL divergence, with cumulative regret reductions of up to 40\% and 50\% when compared to Thompson sampling and Bayes-UCB, respectively.

We argue that the comparative performance loss of Bayes-UCB in the contextual Gaussian case relates to the $\alpha_t = 1/t$ quantile value proposed by \cite{ip-Kaufmann2012}. Its justification comes from the bounds established for the Bernoulli bandit case, but there are no guarantees provided for other bandits. That is, the optimal quantile values for Bayes-UCB are problem dependent, and require careful analytical derivations. On the contrary, our proposed double sampling algorithm does not require any manual tuning, as it autonomously balances the exploration-exploitation tradeoff by adjusting the number of candidate arm samples $N_{t+1}$ based on the learning uncertainty.

\section{Conclusion}
\label{sec:conclusion}

We have presented a new sampling-based probability matching technique for the multi-armed bandit setting. We formulated the problem as a Bayesian sequential learning one, and leveraged random sampling to overcome two of its main challenges: approximating the analytically unsolvable integrals, and automatically balancing the exploration-exploitation tradeoff. We empirically show that additional sampling from the model, which is in many application domains inexpensive in comparison with interacting with the world, can provide improved statistics and, ultimately, reduced regrets. Encouraged by these findings, we aim at implementing this technique with other reward distributions and extending it to real application datasets.

\subsection{Software and Data}
The implementation of the proposed method is available in \href{https://github.com/iurteaga/bandits}{this public repository}. It contains all the software required for replication of the findings of this study.

\subsubsection*{Acknowledgments}
This research was supported in part by NSF grant SCH-1344668. We thank Hang Su and Edward Peng Yu for their feedback and comments on this work.

\bibliography{../literature}
\bibliographystyle{abbrvnat}

\end{document}